# ISIS at its apogee: the Arabic discourse on Twitter and what we can learn from that about ISIS support and Foreign Fighters

*Andrea Ceron*
*Università degli Studi di Milano*

*Luigi Curini*
*Università degli Studi di Milano*

*Department of Social and Political Sciences*
*Via Conservatorio, 7 - 20122 Milano (Italy)*

*Stefano M. Iacus*
*Department of Economics, Management e Quantitative Methods*
*Università degli Studi di Milano*
*Via Conservatorio, 7 - 20122 Milano (Italy)*

**Abstract:** *We analyze 26.2 million comments published in Arabic language on Twitter, from July 2014 to January 2015, when ISIS' strength reached its peak and the group was prominently expanding the territorial area under its control. By doing that, we are able to measure the share of support and aversion toward the Islamic State within the online Arab communities. We then investigate two specific topics. First, by exploiting the time-granularity of the tweets, we link the opinions with daily events to understand the main determinants of the changing trend in support toward ISIS. Second, by taking advantage of the geographical locations of tweets, we explore the relationship between online opinions across countries and the number of foreign fighters joining ISIS.*

The rise and fall of the so-called Islamic State of Iraq and Syria (ISIS) certainly represents one of the most salient political topics over these last few years (Byman 2016). Just to give an idea, even during the campaign for the 2016 US Presidential elections, Donald Trump repeatedly referred to the need of a new counterterrorism strategy against ISIS and promised to "defeat the ideology of radical Islamic terrorism" (Brands and Feaver 2017: 28).

Due to its fast territorial expansion, to the ostentation of cruelty against prisoners and war victims (Kraidy 2017), but also to its innovative communication skills (Farwell 2014) and, since 2015, to its strategy of frequent "*hand-made*" terrorist attacks in Western countries, the ISIS repeatedly grasped the media attention.

In the past, terrorist groups usually relied on traditional mass media to spread their message, let us think for instance to Al Qaeda, which addressed the public by sending declarations recorded on video tapes to Al-Jazeera (Klausen 2015). Conversely, the communication strategy adopted by ISIS was rather different. It has been argued that ISIS was the first Islamic terrorist group that made a massive usage of Internet and it used social networking sites to spread its message in order to generate support (Klausen 2015; Novenario 2016) but also for proselytism (Greenberg 2016).

In this regard, the debate around ISIS propaganda on social media, which started in 2014, was one of the first fire alarms related to the potential perils of social media, linked with the idea of the existence of a "dark web" of online crimes and violence. While, since the Arab spring, academic studies were in fact mainly focused on the potential positive effects of social media in terms of democratization (e.g. Howard and Hussain 2011), the rise of ISIS (also online) questioned previous theories suggesting that social media can also produce turmoil and allow some political actors to pursue illiberal goals (Tucker et al. 2017).

For all these reasons, and under the idea that ISIS has been particularly active online to generate consent and raise followers, it is worth investigating social media conversations about ISIS in order to evaluate the degree of support expressed within the online Arabic communities, to inspect what elements are able to affect its support and to link online opinions with offline outcomes, such as the ability to successfully perform proselytism recruiting foreign fighters.

Using a supervised aggregated sentiment analysis approach, we analyzed 26.2 million comments published in Arabic language on Twitter, from July 2014 to January 2015, when ISIS' strength reached its peak and the group was prominently expanding the territorial area under its control. By doing that, we were able to measure the share of support and aversion toward the Islamic State within the online Arab speaking communities. The fact that the language used by pro-ISIS accounts on Twitter is by far the Arabic one (in over the 90% of the cases; see Siegel and Tucker 2017: 6) makes the choice to focus in the present paper on posts written precisely in Arabic language particularly interesting. By applying statistical analysis to the results of sentiment analysis we investigate two specific topics concerning the conversations on ISIS on-line. First, by exploiting the time-granularity of the tweets, we link the opinions with daily events to understand the main determinants of the changing trend in support toward ISIS. Second, by taking advantage of the geographical locations of tweets, we explore the relationship between online opinions across countries and offline behavior, linking on-line sentiment (i.e., the ratio of positive comments about ISIS: see more below) with data about the number of foreign fighters joining ISIS from those countries.

With respect to the first topic, our analysis show that the expressed on-line support towards ISIS mainly changes according to the specific target of ISIS actions, to military events, to the on-line volume of the discussion about ISIS and to the coverage of media about it. Second, our analysis shows the existence of a robust and negative relationship between the sentiment towards ISIS and the number of foreign fighters in a given country. In this sense, our results appear to unveil the existence of what we have called a "loneliness effect" that could make the "exit" option of becoming a foreign fighter more attractive for some given on-line users. The policy implications of this are far from being trivial: as we will discuss, this finding seems to suggest that censorship is not a solution to counter the ISIS threat: quite the contrary, by decreasing expressed support for the terrorist group, censorship can favor radicalization.

This paper is organized as follows. In Section 1, we review the existing literature linking social media and sensitive topics (with an eye to terrorism) and we outline our main research questions and expectations. In Section 2 we present the technique of sentiment analysis that has been used to catch online opinions. In Section 3 we provide more details on data collection, illustrating our strategy to monitor the discussion about ISIS written in Arabic language. Section 4 and 5 describe the results of our statistical analyses with respect to the two topics highlighted above. A conclusion follows.

1. **Social media and terrorism: literature and research questions**

One of the attractive feature of using big data is that they can allow to observe theoretically relevant social and political attitudes that is normally difficult (impossible?) to detect, what Nagler and Tucker (2015) call the "unfiltered" opinions of individuals.

This is the reason why social media texts have been widely used to study highly sensitive topics such as drug use, sexual behaviors, criminal behavior and controversial political and social issues (e.g. Burnap and Williams 2015; Zeitzoff et al. 2015). To research directly such attitudes and preferences is extremely difficult (Monroe et al. 2015); however, more indirect approaches can be fruitful. As argued in Jamal et al. (2015), contemporary social media enable individuals to express their views in public in relatively safe ways producing as a consequence a set of discourses, possibly not deeply reflective, but still revealing about values, perspectives, and emotions of large numbers of people who have politically relevant views and are ready to express them (at least on-line). For instance, Berinsky (1999) showed that some individuals who harbor anti-integrationist sentiments are likely to hide their socially unacceptable opinions behind a "don't know" response. Under these circumstances, aggregate public opinion may be a poor reflection of collective public sentiment. Bishop (2003) found comparable results with respect to opinion polls conducted on divisive policy issues, such as the teaching of creationism and intelligent design in American public schools. Similarly, Stephens-Davidowitz (2013) by examining Google searches conducted during the 2008 US presidential race found that some US states were more likely to use racial epithets in conjunction with searches on Barack Obama's name, patterns that were not detected by standard survey techniques. Additionally, social media data is less likely to be affected by social desirability bias than polling data (Fisher 1993; DiGrazia et al. 2013).

A growing number of studies has analyzed the relationship between social media and political sensitive topics (including terrorism) also in the context of Arab communities. For instance, Zeitzoff et al. (2015) used a network analysis to examine how foreign policy discussions about Israel–Iran are structured across different languages, including the Arabic one. Zeitzoff (2011) developed an hourly dyadic conflict intensity scores by drawing Twitter and other social media sources during the Gaza Conflict (2008–2009). Al-Rawi (2017), analyzed Facebook posts published in Syria in the aftermath of the Arab Spring to understand the on-line sentiment toward the regime of Bashar Assad. Jamal et. al. (2015) investigated the attitudes expressed in Arabic on Twitter towards the United States and Iran, and found that anti-Americanism is pervasive and intense, but they also suggest that this animus is directed less toward American society than toward the impingement of the United States on other countries.

The literature linking social media and terrorism has been boosted by the rise of ISIS (e.g., Farwell 2014) as this group made a large use of social media for propaganda and proselytism, but also for leisure activities and interpersonal communication (Greenberg 2016; Klausen 2015; Novenario 2016). In this regard, two main streams of research can be detected. One is related to the spread of propaganda and sympathy for ISIS, and the other with the actual effectiveness of online recruiting.

For instance, Mitts (2017) investigates whether the intensity of anti-Muslim hostility in four European countries (France, the United Kingdom, Germany, and Belgium) is linked to pro-ISIS radicalization on Twitter. This analysis shows that local-level measures of anti-Muslim animosity correlate significantly and substantively with indicators of online pro-ISIS radicalization. Siegel and Tucker (2017), via a dataset of over 70 million tweets including tweets containing pro or anti-ISIS keywords between February 2015 and April 2016 (therefore, after the period covered in the present analysis), investigates how successful is the Islamic State's online strategy and to what extent does the organization achieve its goals of attracting a global audience, broadcasting its military successes, and marketing the Caliphate. Similarly, Badawy and Ferrara (2018) explore how ISIS makes use of social media to spread its propaganda and recruit militants from the Arab world and across the globe using a dataset of over 1.9 million messages posted on Twitter by about 25,000 ISIS sympathizers (on this point see also Ferrara 2017; Ferrara et al. 2016). Klausen (2015) analyzed approximately around 30,000 Twitter accounts linked with ISIS and discovered that propaganda flew from accounts belonging to terrorist organizations in the insurgency zone, to ISIS sympathizers in Western countries. Indeed, ISIS employed social media (Twitter, Facebook and Instagram) to influence not only friends but also rivals and journalists and, in order to build support, ISIS distributed emotional messages depicting its members as fearsome warriors and claiming that ISIS victory was inevitable (Farwell 2014).

However, "social media is a double-edged sword" for ISIS in terms of support (Farwell 2014: 52). The ostentation of atrocities committed against prisoners and war victims (Kraidy 2017) can generate a backlash effect: ISIS communication can be used by opponent groups to discredit the terrorists and mobilize criticism, reducing ISIS level of online support (Farwell 2014). This could be particularly true when the victims are other Muslims. As is well-known, when a terrorist group chooses a target that is viewed as illegitimate by its constituents, the group can suffer a significant loss in terms of popular respect, trust, and support (Cronin 2009). Indeed, "during an earlier phase of conflict in Iraq, al-Qaeda realized that images of Muslims killing Muslims were counterproductive, and became critical of ISIS for carrying out such actions" (Farwell 2014: 52).

This opens questions about the extent to which social media communication is able to build support for terrorism and suggests that different forms of communication, but also different strategies, behaviors and offline events can be more or less effective in fostering support. Accordingly, we formulate our first research question.

*Research Question 1 (RQ1): What elements influence the level of online support within Twitter Arabic communities?*

A second stream of research pays attention to social media activities devoted to recruitment and discusses counterterrorism strategies focusing on the idea of shutting down ISIS accounts on Twitter and Facebook. In this regard, some studies support the idea of taking ISIS sites down and banning their social media accounts (Cohen 2015; Greenberg 2016) as a strategy to reduce recruitment and "push the remaining rank and file into the digital equivalent of a remote cave" (Cohen 2015). Similary, others argue that "getting ISIS off of popular platforms diminishes their reach and their effectiveness". If ISIS activists move to "dark web" platforms, people could in fact be less likely to actually join ISIS as the effect of propaganda in such low-populated shadow sites should will be lower compared to widely-used social networking sites where not-yet-indoctrinated users proliferate (Greenberg 2016: 176). The fact that a relative large proportion of Western recruits have been shown to have consumed extremist content on the Internet and social media strengthens this conclusion (Carter et al. 2014).

Although this reasoning seems straightforward, censorship can produce unintended results and "relying on the internet exclusively, or even too heavily, can have negative consequences" (Greenberg 2016: 175). Firstly, some scholars are worried because the open web is a source of information for intelligence services (Akhgar et al. 2016). Analyzing public comments posted on social media is a promising way for dealing with terroristic propaganda online (Neumann 2013), whereas information shared on dark web platforms might no longer be accessible nor shut down and allows to protect the identity of ISIS supporters (Cox 2015). Secondly, pushing ISIS supporters into a "digital cave" can actually transform this dark web in a stronger echo-chamber that reinforces radicalism and promote violent extremisms (e.g., Wojcieszak 2010; for a review: O'Hara and Stevens 2015) increasing support for a pan-Islamic project (el-Nawawy and Khamis 2009), like the one initially proposed by ISIS.

Thirdly, and most relevant for the present study as we will discuss below, the literature on terrorism suggests that political violence is a substitute for nonviolent expression of harsh dissent (Frey and Luechinger 2003; Gurr 2006; Lichbach 1987), which can be denied by shutting down ISIS accounts. Sympathizers only surrounded by anti-ISIS voices (an echo-chamber of a different kind), might feel isolated (a "*lonely wolf effect*"); in turn, the unavailability of a "voice" option could bring them to opt for the "exit", entering into pro-ISIS echo-chambers and radicalizing their views to suddenly join ISIS.

This leads us to formulate our second research question:

*Research Question 2 (RQ2): What is the relationship between online support for ISIS and the number of foreign fighters and how social media censorship might affect such relationship?*

2. **From texts to information: our approach**

To investigate the online opinions toward ISIS in this paper we adopt the technique iSA (integrated Sentiment Analysis: Ceron et al. 2016), derived from the fundamental work of Hopkins and King (2010). iSA is a *supervised* and *aggregate* alghoritm. The idea of supervised learning is rather simple (Grimmer and Stewart 2013): it starts with acknowledging the fact that human coders are better in tagging into a predetermined set of categories a set of documents than any automatic approach (via a dictionary for example). The algorithm then "learns" how to sort the documents into categories using the training set and words. The now trained, or supervised, statistical model is then employed to classify the remaining (unread) documents.

The supervised methods aim either to classify the individual documents into categories, via machine learning algorithms, or measuring the proportion of documents in each category, as iSA does (Grimmer and Brandon 2013). The choice between which of the two approaches to adopt is driven by both theoretical (and statistical) reasons. Regarding the former aspect, if the main aim of the research is, as in the present case, to focus on some aggregate generalizations about populations of objects (in our case: the percentage of support towards ISIS), then the task of quantification (i.e., estimating category percentages) is more coherent to it than the task of classification (i.e., classyfying individual documents).

This is also strengthened by two further statistical points: a) shifting focus to estimate directly proportions, rather than doing individual classification and only after aggregate such individual classifications to retrieve for example information about the aggregate sentiment, can lead to substantial improvements in accuracy on the final results; b) no statistical property must be satisfied by the training set for this approach to work properly: that is, the training set is not required to be a representative sample of the population of texts, as it happens with machine learning algorithms.[1] Relaxing such assumption allows to dramatically reduce (by more than 20 times: Ceron *et al.* 2016) the required size of the training-set needed with respect to a given test-set in order for the analysis to produce reliable estimates (Hopkins and King 2010).

---

[1] The only assumption to be satisfied in the "aggregate" approach is that the language used in the training-set to express some given concept must be the same as in the whole population of posts, i.e. social media users use the same language (Ceron et al. 2016).

The statistical properties of iSA alone and in comparison with other machine learning tools have been shown in full details in Ceron *et al.* (2016) to which the reader is redirected.[2]

### 3. Data collection and preliminary results

In our analysis, we focused on Twitter data. As a social network, Twitter is popular within the Arabic community. Nearly 40% of the Arab public is now online and of this population, 30% are on Twitter (Jamal et al. 2015). Of course, social media and similar data reflect only the population from which they were extracted, as well as the specific topic this population is debating about in a given time period (DiGrazia et al. 2013; Nagler and Tucker 2015). It remains however undisputable that social media debates affect participants' expectations about how other on-line participants will respond to their own posts (Jamal et al. 2015). They are therefore likely to affect participants' own expressions of views through persuasion and socialization and by shaping their incentives with respect to their own contributions. Hence, these discourses are politically important in their own right. Considering the percentage of people using Twitter within the Arab public just strengthens this point. Moreover, Twitter has been shown to be used repeatedly by ISIS as a propaganda tool (Berger and Morgan 2015), given the technical advantages provided by this micro-blogging social media such as large-scale public dissemination of content (Klausen, 2015).Therefore it becomes natural to focus on it.[3]

As already noted, we decided to focus on a specific time-period: the one between the 1st of July 2014 and the end of January 2015. There are three main reasons underlying such choice. First this period represents the moment in which Isis was at its apogee in terms of territorial expansion with the conquest of Sinjar (August 2014) in Iraq and symbolic power as the self-proclaimed Caliph Abu Bakr al-Baghdadi made its first (and last) public experience in a mosque of Mosul on the 5th of July. During this period, ISIS was at the center of the media discourse of many Arab states, given that the International Coalitions, which has included some of those countries (Jordan, Morocco, Bahrain,

---

[2] The R package names iSAX implements iSA technology and it is freely available at: https://github.com/blogsvoices/iSAX

[3] The fact that Twitter data is more easily accessible contrary to other social networks (such as Facebook, Instagram) is of course the other (often untold) reason of why so many researches in these last years have focused on it rather than on some other alternatives.

Jordan, Qatar, Saudi Arabia, and the United Arab Emirates) started to target the Islamic State since the 8th of August 2014 in Iraq and the 22nd of September in Syria. Second, although the suspension activity puts forward by Twitter itself against ISIS-supporting accounts started already in late 2014, its peak happened only several months later.[4] As a result of that, the period we covered in our analysis should be one in which people could feel less the pressure to reveal his/her true opinion towards Isis without the risk of being censored for that. This is, after all, the reason why focusing on social media is interesting, as discussed above. Third, and finally, by extending the analysis till January 2015, we are able to capture the impact on the debate about Isis within Arab on-line community of the major terrorist attack committed by a group of ISIS inspired terrorists in Paris (i.e., Charlie Hebdo) that had a world-wide echo.

For the present paper, Twitter data has been collected via Brandwatch, an official firehose company dealing with Twitter, on all Arabic language tweets that explicitly discussed ISIS (see the Appendix for the list of keywords employed in our query). By relying on a firehose, and therefore by collecting the entire universe of tweets that satisfy our search query, our aim was to avoid the possible sampling bias introduced in the study of Twitter when collecting data through publicly available APIs (González-Bailón et al. 2014).[5] Our final number of tweets was 26.2 million (on average 128,000 per day; median value: 99,000).

---

[4] Berger and Morgan (2015), for example, show that the original network of Isis activists suffered only a 3.4 percent loss in membership between September 2014 and January 2015. On the contrary, in the following months up to early 2016 Twitter suspended around 125,000 accounts delivering pro-ISIS messages (Siegel and Tucker 2017).

[5] The Arabic language has many varieties and dialects. This brings another complexity when dealing with sentiment analysis, since most Arabic users express their opinion using local dialects instead of Standard Arabic (Al-Moslmi et al. 2017). We focus on the posts written in Modern Standard Arabic besides the Levantine and Egyptian dialects, classifying as off-topics all other dialects. Note also that by deciding to focus only on tweets written in Arabic language, we could have incorporated a bias in our analysis as long as the subset of Twitter users in the social-media Arabic sphere who employs other languages to debate about ISIS shares some peculiar and unique characteristics with respect to our research focus (i.e., detecting the sentiment towards ISIS). This is however a limit that couldn't be easily avoided given the geographical scope of our analysis that covers a large number of countries, and therefore of different languages (other than the Arabic one).

Applying sentiment analysis on Arabic posts has attracted a growing interest in recent years (e.g., Al-Moslmi et al. 2017). As discussed in the previous section, we relied on iSA to discern the position expressed by users towards ISIS. For doing that, in the training-set stage of the analysis, we employed three graduated Arabic native speaker students (one Syrian, one Egyptian and one from Morocco) to ascribe the tone (that we call "sentiment") towards ISIS with 3 options: positive, negative or neutral.[6] The training-set was compromised by 1,600 tweets extracted randomly from different days in the period here analyzed.[7] See Table 1 for some examples of tweets and their sentiment classification (expressing either a positive or a negative attitude towards ISIS).

Table 1 here

Overall, the average positive sentiment value towards Isis (the ratio between % of positive tweets over the sum of % of positive and negative tweets) is 25.1%. In the already quoted work by Jamal et. al. (2015), the authors investigate the reactions to terrorist events (such as the Boston Marathon bombing in 2013 and an attack in London) within the Arabic twitter-sphere by employing the algorithm developed by Hopkins and King (2010), that, as already highlighted, shares the same aggregated and supervised approach as iSA. Quite interestingly, they find that the degree of explicit support in each of the two above mentioned terrorist attacks was roughly one comment out of four, among those who took a clear and strong position on Twitter, therefore producing a result quite similar to ours. Although their result has been derived by focusing on a different subject in an earlier period than ours, we take this recurring percentage (25%) as an indirect corroboration of our recovered aggregate measure of (positive) sentiment towards ISIS in the Arabic Twitter-sphere.

Of course an average value could mask a large variance in the data. Figure 1 in this regard focuses on the daily variation around the share of positive sentiment mean towards ISIS's actions, a variable we label SENTIMENT DEVIATION. Any value higher than 0 means therefore that in that day the expressed (positive) sentiment towards Isis was higher than the mean. The opposite for any value lower than 0. In the same graph we have also reported the corresponding lowess function. Interestingly, this lowess function presents a rather flat trend, meaning than any extemporal shock in the value of SENTIMENT DEVIATION appears to be absorbed in a very fast rate. However, the lowess function begins to bend down in January 2015.

---

[6] In our analysis, we do not distinguish between original tweets and re-tweets, so our implicit assumption is that re-tweets usually reflect sympathy with the original tweet. The inter-coder reliability on the positive and negative categories among the coders on a sub-sample of 200 tweets is 0.93 (percent agreement).

[7] In our analysis we did not apply any stemming procedure to Arabic texts, given the still controversial debate on the topic. See however the stemming package 'arabicStemR', recently developed for R users by Richard Nielsen (Nielsen 2013).

Figure 1 here

Similarly, in Figure 2 we have plotted the daily variation around the average number of all tweets (including the neutral ones), a variable that is labelled ATTENTION DEVIATION. In this case, a positive number means on that day the volume of discussion about ISIS within the Arabic Twitter-sphere was larger than the mean (around 128,000 tweets, as noted above); a negative number indicates the opposite. Also in this Figure we have plotted a lowess function that shows a substantial stability that begins to increase once again around January 2015.

Figure 2 here

## 4. Determinants of daily positive sentiment towards ISIS

In our first econometric analysis, we aim to explain the temporal variation in the data recorded for the SENTIMENT DEVIATION variable. For doing that, we focus on a set of variables related to several major events.[8]

First we control for the *Beheadings* committed by ISIS and often broad banded via social media. In this respect, we differentiated between beheadings committed against Western (such as the American journalist James Wright Foley in August 2014) as well as Non-Western victims (such as the Lebanese Army soldier Abbas Medlej in September 2014).[9] Then we focus on the *military events* characterizing the on-going military campaign by ISIS in Syria and Iraq, differentiating in this case between major victories (such as the seized by ISIS of Syria's Shaer gas field in Homs Governorate happened the 17 July 2014) and major defeats (such as the Battle of Suq al Ghazi that ended with a US–Iraqi win the 15 September 2014). We also control for the "object" of a ISIS attack during the period here considered, in particular by focusing on those *attacks against Mosques or Imams* (such as the killing of the Sunni Imam Abdul Rahman al-Jobouri in Baquba the 22 July 2014). Finally, we include those days in which major ISIS actors gave some *speeches calling for "Muslism Unity"* (such as the speech pronounced the 21st September 2014 by ISIS Official spokesman Abu Mohammad al-Adnani that encouraged Muslims around the world to kill non-Muslims). Finally, we include in our analysis the lag of POSITIVE DEVIATION to account for the data's temporal structure. The results are reported in Table 2.

---

[8] This also means that we dismiss small, local events that might well impact individuals in close proximity, but that should not necessarily affect users in different countries.

[9] Source: https://en.wikipedia.org/wiki/Timeline_of_ISIL-related_events

Table 2 here

Model 1 shows several interesting findings. First, it appears a kind of "us vs. them" effect: the *attacks against Mosques or Imams* variable shows in fact a negative impact on SENTIMENT DEVIATION confirming a well-known finding in the terrorist literature (Cronin 2009). Indeed, terrorist groups gain their legitimacy by claiming that they are acting on behalf of a larger cause, however, targeting errors can lead to a significant loss of popular respect, trust, and support. In this sense, when a terrorist group chooses a target that is viewed as illegitimate by their constituents (such as precisely attacking Mosques or Imams in case of ISIS), the group can lose a significant amount of popular support, at least temporary. .[10]

Also "war events" appear to matter: in particular, *ISIS Military victory* decreases SENTIMENT DEVIATION as a possible result of an increased "fear effect" within Arabic communities. Other forms of violence do not seem to effect systematically the sentiment towards ISIS: in this sense, we do not find any evidence of a clear backlash effect related to the ostentation of atrocities committed against prisoners often highlighted in the literature, although it is worth stressing the negative sign found for the beheadings of Non-Western prisoners compared to Western ones. A significant and once again negative impact on SENTIMENT DEVIATION is also played by the ISIS call for unity of Muslim. This is not surprising, once we recognize how the main reason to criticize ISIS in the Arabic Twitter-sphere results the use by ISIS of Islamic religion as a shield for pursuing political aims (size power and rule a State), that is perceived to be solely for private interest (e.g., Malik 2014). Finally, the fact that the lagged dependent variable is not significant confirms what we have visually already noted earlier: any extemporal shock on SENTIMENT DEVIATION does not appear to produce any gradually adjusting dynamic. On the contrary, such shock is immediately absorbed in the trend of SENTIMENT DEVIATION, implying probably that any substantial change in SENTIMENT DEVIATION is just due to the entrance (or the exit) in the Arabic Twitter-sphere of new users that discusses about ISIS, rather than any change of opinions of those users that usually discuss about such topic (more on this below). In Model 2 we show the role played by the degree of attention, in terms of number of posts, devoted to discuss ISIS within Arabic Twitter on the sentiment towards ISIS. As already noted in Figure 2, since January 2015 the volume of discussion about ISIS experiences a sharp increase. One of the reason for this outcome is clearly related to what happened between the 7[th] and the 9[th] of January in Paris, i.e., the attack against Charlie Hebdo committed by a group of ISIS inspired terrorists that had a world-wide echo.[11] We therefore introduced a dummy equals to 1 for the *Charlie Hebdo attack*. Moreover, we also included as a further control variable the ATTENTION DEVIATION discussed in Figure 2. As can be seen, both variables are highly significant and negative correlated with SENTIMENT DEVIATION. In particular, with respect to ATTENTION DEVIATION, an increase by 100,000 posts with respect to the average number of posts devoted to ISIS in normal time, produced a contraction in SENTIMENT DEVIATION equals to 1 standard deviation of its value. This can also be appreciated by noting that if, as already noted, the average value of positive sentiment towards ISIS in our dataset is 25.1%, this value drops to 23.6% once we weight the average according to the number of tweets, and to 12.9% in those days in which we have more than 200,000 comments (90% percentile). This is coherent with what found in Stigler and Tucker (2017). In their paper, between February 2015 (therefore after *Charlie Hebdo attack*) and April 2016, the estimated sentiment pro-ISIS is 13.9% given, however, an average daily number of tweets higher than the one we have registered in the present paper (on average around 160,000). This also implies that during "normal time" of on-line attention, we probably oversample pro-ISIS supporters, while during "exceptional time" of on-line attention, a larger share of public that normally does not discuss about ISIS, enters in the on-line debate, expressing a sentiment that on average is largely negative towards ISIS.

---

[10] On this same conclusion, see also Cunliffe and Curini (2018).

In Model 3, on the other hand, we controlled how the number of articles published on Arabic on-line newspapers about ISIS affects the sentiment recorded on Twitter. We employed the same keywords employed to monitor the Twitter activity (see Appendix) to recover the on-line news written in Arabic language that were discussing about ISIS in the temporal period covered in our analysis. Also in this case, the data comes from Brandwatch. The average number of articles discussing about ISIS on a daily basis is 737. In this respect, we found a positive media effect (the more news on-line discusses about ISIS, the higher is SENTIMENT DEVIATION). An extended literature shows how heavy media coverage of terrorist activities can increase the likelihood that similar actions will occur in the future (see Jenkins 1981; Weimann and Winn 1994). The results reported in Model 3 show that there is the possibility of a positive effect of media attention not only with respect to terrorist actions, but also on the support expressed on-line toward such groups, the more these groups are frequently discussed by journalists.

---

[11] At least one of the person involved in the terrorist attacks between 7th and 9th January, Amedy Coulibaly, had pledged allegiance to ISIS.

### 5. Positive Sentiment towards ISIS and Foreign Fighters

The phenomenon of foreign fighters, i.e., people who decide to leave their own country to go to fight for ISIS in Syria and Iraq, has attracted, as discussed earlier, a large attention in these last years, both at the academic level as well on the popular press as already discussed above. A special focus has been devoted in particular on the reasons that could explain such radical choice. In this section we exploit the geo-localization of tweets to understand if there is any the relationship between the national on-line overall tone towards ISIS across countries with the number of foreign fighters for ISIS of those same countries.

To identify the national origin of a tweet we followed these rules: a) we considered the geo-coordinates meta-data attached to a tweet whenever they were available; b) otherwise, to determine the location of a mention we took advantages either of the information provided directly by the user and/or the time zone meta-data that is sometimes attached to a tweet. Through this method, we were able to recover the national origin of 45% of tweets in our dataset.[12] Note that we included in the analysis reported below only those countries that present more than 1,000 tweets estimated according to the procedure just discussed. This allows us to have a reasonable amount of data to analyze from every country. The final sample of countries in the analysis reported below is 61.

The data source for foreign fighters comes from a study by International Center for the study of Radicalisation and Political violence published in January 2015.[13]

Behind the *Sentiment toward ISIS* variable in a given country (estimated as usual as the percentage of positive statements about ISIS over the sum of the percentage of positive and negative statements about ISIS), we control for several other variables at the country-level: *Active Islamic Terrorist Group* is a dummy equals 1 if there is one or more Islamic terrorist group active within a given country;[14] *Living in a country around ISIS border* is once again a dummy equals 1 if a country shares a border with the ISIS state; *% Shia over Muslims within a country*;[15] *% Broadband* that we treat as a proxy for fast-Internet diffusion;[16] finally, we also include the *Democracy score* for a country according to Polity IV.

---

[12] An alternative method to recover the geo-localization of tweets is to to triangulate a user's location based on locations provided by their networks of friends and followers via a spatial label propagation algorithm (see Jurgens 2013). For an application to the analysis of social media data related to ISIS, see Mitts (2017).

[13] Source: http://icsr.info/2015/01/foreign-fighter-total-syriairaq-now-exceeds-20000-surpasses-afghanistan-conflict-1980s/

Given the characteristics of our dependent variable, we estimate a set of Negative Binomial models (see Table 3). Model 1 is our benchmark model where we include all countries presented in our dataset. Models 2 to 5 allow to control for the robustness of our findings. Let's start to discuss the results from Model 1.

Table 3 here

As can be seen, *Sentiment toward ISIS* is always significant and negative. That is, when a given country has a higher ratio of positive to negative ISIS-related tweets, fewer of its residents traveled to fight with the Islamic State. Figure 3 shows in this respect the predicted number of Foreign Fighters according to *Sentiment toward ISIS* (source: Model 1 in Table 3). So, for example, the expected number of Foreign Fighters decreases by more than half if the *Twitter-Sentiment toward ISIS* in a country increases from 10% to 20%.

Figure 3 here

The opposite is also true: the more negative a country's Twitter discussion was about the Islamic State, the more of that country's people left to fight with the group.

Why this result? The exit and voice seminal framework originally proposed by Albert Hirschman (1970) is useful in this regard. As is well known, Hirschman underlines how any member of an organization, whether a nation, a private business or any other form of human grouping, have essentially two possible responses when she perceives a change in her social environment connected to some actions (or inactions) put forward by her organization: she can exit (withdraw from the

---

[14] Source: https://en.wikipedia.org/wiki/List_of_designated_terrorist_groups

[15] Sources: http://www.pewforum.org/datasets/the-worlds-muslims/;
http://en.wikipedia.org/wiki/Shia_Islam.

[16] Source: Worldbank database.

- 15 -

relationship); or, she can voice (attempt to repair or improve the relationship through communication of the complaint, grievance or proposal for change).

Following exactly this same theoretical framework, our hypothesis is that when Islamic State sympathizers find (or at least perceive according to the on-line debate and the connected level of *Sentiment toward ISIS* therein arising) an environment in their respective countries where they can share and discuss their ideas - even extremist ones - fewer of them could feel the need to take action by leaving "home" to go fight for the group. The opposite happens in a context in which they feel to be isolated in terms of preferences (i.e., with a low level of *Sentiment toward ISIS*). Using Hirschman's jargon, individuals struggling to be "loyal" to a community with non-radical preferences, instead of using their "voice" (or, more pertinently, their tweet) to express their radical preferences, could rather decide to opt for a far more radical "exit" and actually join the Islamic State.[17] The existence of a "loneliness effect" could make, in other words, this latter option more attractive for at least some people.[18] This explanation is also coherent with a relevant literature that although focusing on violent activity rather than looking directly at recruitment (as in the case of Foreign Fighters), stresses how the tolerance of radical voices could have positive feedback. For example, Lichbach (1987) shows that nonviolent action is substituted for violent action when the violent action is repressed, and vice-verse. Similarly Frey and Luechinger (2003) show that decreasing opportunity costs for nonviolent activities decreases use of terrorist violence, while Gurr (2006) stresses that allowing nonviolent expression of dissent lowers incentives for engaging in violent action.

---

[17] A possible critique in this regard is that, by definition, online communities are not bounded by any national borders. So what should really matter in the exit-voice dynamic discussed above is, eventually, the level of *Sentiment toward ISIS* arising in the entire internet, not in one's own "national on-line community". Two answers are available to this critique. First, as highlighted, the *Sentiment toward ISIS* in one's own country is just a way for a ISIS sympathizer to perceive how people are supportive (or not) about his/her ideas around him/her in one's own community. Second, while social media (including Twitter) allow people to connect with others across the globe, recent studies have found that physical relationships in the offline world still strongly influence online social relationships (see Mitts 2017; Jurgens 2013). As a result, a large share of individuals' online social media usually includes geographically close friends.

[18] Also Mitts (2017) highlights how the experiences of social isolation can trigger individuals to open up to extremist jihadi ideologies. She however focuses on a different source of social isolation than ours (local exposure to anti-Muslim hostility in four European countries).

We ran a set of further models to check for the robustness of our finding. In Model 2 we replicate Model 1 but including this time only those countries with a number of tweets discussing about ISIS in Arabic language larger than 15,000. Model 3 drops from the analysis the United States. United States is in fact the country with the largest number of tweets (almost 2/3 of the total) and, as already noted (Berger and Morgan 2015), there are several reasons to treat with extra care the data geolocated in the United States when the discussion about ISIS is concerned. As can be seen, in both cases, *Sentiment toward ISIS* remains significant and with the usual negative sign.[19]

There could also be some alternative explanations behind the negative relationship that we found between *Sentiment toward ISIS* and the number of Foreign Fighters. *First*, it could be argued that it is precisely because a large number of residents has left a given country to become foreign fighters during the temporal period we cover in our analysis (July 2014-January 2015) that explains such a negative relationship. The reasoning here is the following one: by leaving country A in those months, those residents stop to publish their pro-ISIS opinions on Twitter from country A, depressing through that the value of *Sentiment toward ISIS* that we retrieved from that same country. To take into consideration such alternative explanation, in Model 4 we use the update data on the number of Foreign Fighters that comes from a different study than the one used till now. This study by The Soufan Group has been published almost 3 years after the temporal period we covered in our analysis (October 2017)[20], therefore it includes also all the foreign fighters that left a country following the time in which *Sentiment toward ISIS* has been estimated. A lower value for this latter variable, therefore, cannot be linked to the alternative process just highlighted above.

*Second*, it could be that in countries where most people have gone to fight for ISIS, attitudes toward ISIS, including on social media, are more negative. If that were the case, a reverse causation between *Sentiment toward ISIS* and our dependent variable would of course appear. To check for this, in Model 5 we use as a proxy for *Sentiment toward ISIS* the answers that come from a 2013 survey of the Pew Research Center titled "The World's Muslims: Religion, Politics and Society".[21] In particular, we used the percentage of respondents who justify suicide attacks in the set of countries included in our database (we label this variable *Justify Attacks*). Only a significant negative effect of *Justify Attacks* on our dependent variable would be coherent with the "loneliness effect" highlighted above.[22]

---

[19] Several ISIS supporters appear in fact to deceptively listed locations in the United States in order to create the appearance of a homeland threat. In a further model, we have also controlled for a variable capturing the volume of tweets in each country as well as for an interaction between this variable and *Sentiment toward ISIS*. Our results hold intact also in these two latter cases.

[20] http://thesoufancenter.org/wp-content/uploads/2017/10/Beyond-the-Caliphate-Foreign-Fighters-and-the-Threat-of-Returnees-TSC-Report-October-2017.pdf

Both Models 4 and 5 are reassuring about the robustness of our explanation linking *Sentiment toward ISIS* and the number of Foreign Fighters. *Sentiment toward ISIS* remains in fact negative and significant (see Model 4) also when using data of Foreign Fighters following January 2015 (when we stopped to analysis posts on Twitters), while in Model 5 the variable *Justify Attacks* is similarly showing a significant and negative impact on our dependent variable.

With respect to the other variables included in Table 3, the fact that the *Democracy score* variable when significant has a negative score provides a further evidence of the dynamics underlined above: a "marketplace of ideas" connected with openness in terms of political participation that are typical of a well-functioning democracy (both on-line as well as off-line) can in fact at least partly prevent radicalization (in our case, the number of foreign fighters leaving a given country) by allowing a debate that can divert potentially violent behavior (Briggs 2010; Dalacoura 2006).

It is also interesting to note how everything that reduces the "transaction costs" of becoming a foreign fighter (such as *Living in a country around ISIS border, % Broadband*, as well as, although with a much weaker effect, having within a country an already existing *Active Islamic Terrorist Group*,: all factors that could make communication, access to information to organize a trip to Syria and Iraq easier as well as reducing the financial costs of such trip), increases almost always the number of foreign fighters from a given country. Finally, the existence of a strong and negative impact of *% Shia over Muslims within a country* on the numbers of foreign fighters highlight the well-known religious divide inside Islam characterizing the Shia-Sunni relationship (Hashemi and Postel 2017). This is far from being a surprise: ISIS rhetoric has often fanned the flames of sectarian hostility by presenting the Caliphate as the defender of Sunnis against both Shia-led militias and governments and defining Shias collectively as infidels and therefore legitimate targets of jihad. (Hassan 2016; Gatenstein et al., 2016)

---

[21] http://www.pewforum.org/2013/04/30/the-worlds-muslims-religion-politics-society-overview/

[22] Given the lower number of observations in Model 5 (22) compared to Model 1 (61) we just included in the model the variable *Justify Attacks*. Note however that replicating Model 5 with the other independent variables included in Model 1 does not affect any of our conclusions.

**Conclusion**

Compared to the time-frame covered in the present analysis, the military campaign against ISIS in both Syria and Iraq has recorded an impressive series of victories in the months following January 2015, forcing ISIS to abandon almost all of the territory previously controlled. This, however, does not reduce the relevance of the results reported in the present study that focuses on the "glorious days" of the so called "Islamic State": not only in terms of better understanding an (almost) past phenomenon, but also in terms of the lessons we can derive from it. After all, despite the current military defeat, concern that ISIS may remain viable in the long term, both as group and as an inspiration, has not faded away (Barrett 2017).

Research suggested, as already underlined, that "social media is a double-edged sword" for ISIS because it allows spreading propaganda but also increases ISIS vulnerability – particularly when the victims are other Muslims, mobilizing the opponents and discrediting the terrorist group (Farwell 2014: 52). Building on this debate, our results about the Arabic Twitter discourses on ISIS suggest – indeed – that the support toward ISIS drops, for example, when the group attacks rival Mosques and opponent Imams, killing other Muslims.

Furthermore, in line with the idea of a lonely wolf, our analysis shows that the number of foreign fighters is lower in countries that report a stronger positive sentiment toward ISIS. Conversely, ISIS sympathizers living in countries with a lower share of support might feel marginalized and end up radicalizing their views joining ISIS.

The policy implication of such finding are not trivial. After the tragic events in Paris, the group Anonymous hacked and shut down 5,000 Twitter accounts held by Islamic State sympathizers, with a great deal of media attention. Twitter itself and various security agencies (including Europol) had already been using similar strategies to limit the group's followers' ability to spread propaganda via social networks. Clearly their main and common aim was to neutralize the Islamic State's ability to use Twitter to reach far beyond its own narrow audience, and to reduce the violent radicals' ability to manipulate public opinion and attract new recruits and sympathizers. The goal might be a good one. But we need to watch out for the unintended - and potentially serious - consequences of such a strategy. Torres Soriano (2012) and Benson (2014), for example, contend that the internet creates as many vulnerabilities for terrorist organizations as it does strengths. On one hand, networked media provides an array of unprecedented advantages for terrorist groups, such as interconnectivity, anonymity, cheapness, power enhancement, access to new audiences (Whine 1999). On the other hand, it also provides a host of disadvantages. Among them, chiefly opportunities for tracing and monitoring, hacking, individual and organizational attacks on sites and disinformation. In this sense, suspending Twitter accounts destroys an important source of intelligence (Akhgar et al. 2016; Neumann 2013).

Second, as much as networked media facilitates communication between like-minded individuals, it also provides a forum for those who are persuasive against it (Torres Soriano 2012).

Third, and finally, limiting debate in a digital forum could further radicalize and isolate possible Islamic State sympathizers. The resulting *loneliness effect* that we find in our data can be dangerous. In this framework, by decreasing expressed support for the terrorist group, censorship can favor radicalization.[23] As a result, to prevent such outcome, it could not be enough for a nation's policies and counterterrorism efforts to welcome minorities into their nation's mainstream, preventing them from feeling marginalized. Quire paradoxically, they could also need to pay attention to individuals within those minorities who begin to feel marginalized within their own moderate communities. A much more difficult result to achieve.

This conclusion, moreover, keeps its relevance also nowadays: ISIS's losses of terrain in the Middle East (and North Africa), seems for example to coincide to a growing online activity by ISIS itself targeting those populations more receptive to its message in several European countries.[24] The fact that ISIS appears successful at inspiring low-level attacks in Europe despite its territorial losses indicates in this sense that its messaging for a "call for lone jihad" remains potentially resonant. Such message, moreover, could be easily interpreted as a different factual translation of the "loneliness effect" we have highlighted in our paper: by being deprived by the exit route represented by joining the Islamic State in the Islamic State controlled territory, the "loneliness effect" could risk to produce a new "(extreme) exit" option, i.e., a suicide terrorist attack in one's own native country. Adding to this scenario the growing number of returning Foreign Fighters (at least 5,600 citizens from 33 countries that have left Syria and Iraq to return back to their own community at the end of 2017: see Barrett 2017) just increases the relevance of what underlined. This represents a future direction of research that should be investigated in greater details.

---

[23] Of course, avoiding censorship on extremist voices, per-se, appears at beast just as a necessary not a sufficient condition to deter the consequences of the loneliness effect. Other factors that could play a relevant role are the developing of an efficient "counter-narrative" focused on delegitimizing extremist ideologies (Knops 2010) as well as social welfare policies directed at grievance alleviation (Burgoon 2006). Mitts (2017) also underlines the role played by political factors in some countries (such as the extent of anti-Muslim rhetoric).

[24] See:
http://www.understandingwar.org/sites/default/files/ISIS%20in%20Europe%20Update%20September-2017.pdf

Our analysis is not without limitations: in particular, by focusing on the aggregate level, as we do in the present paper, we cannot rule out the possible existence of an ecological fallacy in our results. Still, as we have shown, the fact that our results appear to be robust to several alternative checks, is reassuring about our conclusion.

Our results sends back to a number of studies that consider political terrorism as a substitute for nonviolent expression of harsh dissent (Frey and Luechinger 2003; Gurr 2006; Lichbach 1987). This can be also of interest to the growing literature discussing about censorship (King et al. 2013, 2014) and the effects of social media on democracy (Tucker et al. 2017). Finally, this study adds to the mounting evidence that online social networks are not ephemeral, spam-ridden sources of information (DiGrazia et al. 2013). Rather, social media activity can provide a valid indicator of political decision making that could have relevant (and sometimes unfortunate) consequences.

DiGrazia J, McKelvey K, Bollen J, Rojas F (2013) More Tweets, More Votes: Social Media as a Quantitative Indicator of Political Behavior. *PLoS ONE* 8(11): e79449. doi:10.1371/journal.pone.0079449

Efron, B., Tibshirani, R. (1993). *An Introduction to the Bootstrap*, Boca Raton, FL, Chapman & Hall/CRC,*CRC* press.

el-Nawawy, M., and S. Khamis (2009). *Islam Dot Com: Contemporary Islamic Discourses in Cyberspace*. New York: Palgrave Macmillan.

Farwell, James P. (2014). The Media Strategy of ISIS. *Survival* 56(6): 49-55

Ferrara, E. (2017). Contagion dynamics of extremist propaganda in social networks. *Information Sciences*, 418, 1-12.

Ferrara E., Wang WQ., Varol O., Flammini A., Galstyan A. (2016) Predicting Online Extremism, Content Adopters, and Interaction Reciprocity. In: Spiro E., Ahn YY. (eds) *Social Informatics*. SocInfo 2016. Lecture Notes in Computer Science, vol 10047. Springer, Cham

Fisher, R.J. (1993). Social desirability bias and the validity of indirect questioning. *Journal of Consumer Research*, 20(2), 303–315.

Frey, Bruno S., and Simon Luechinger (2003). How to Fight Terrorism: Alternatives to Deterrence. *Defense and Peace Economics* 14(4): 237-249.

Gartenstein-Ross, D., Barr, N. and B. Moreng. (2016). "The Islamic State's Global Propaganda Strategy", *The International Centre for Counter-Terrorism – The Hague* 7, no. 1 . [Online]. 31st March. Available from: https://icct.nl/publication/the-islamic-states-global-propaganda-strategy/

González-Bailón, Sandra, Wang, Ning, Rivero, Alejandro, Borge-Holthoeferd, Javier, and Moreno, Yamir (2014). Assessing the bias in samples of large online networks, *Social Networks*, 38, 16-27

- 24 -

**Figures**

*Figure 1: Daily variation around the share of positive sentiment mean towards ISIS's actions (variable name: SENTIMENT DEVIATION)*

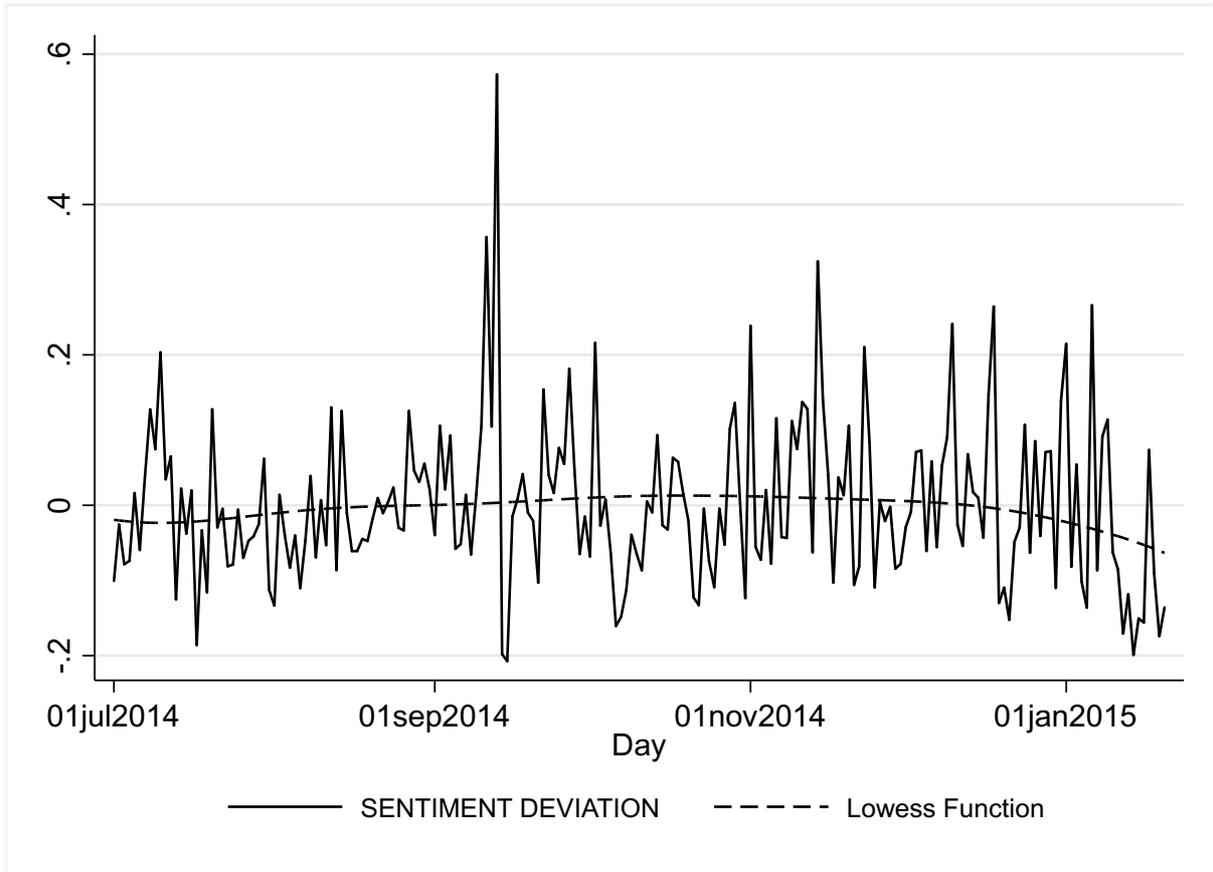

*Figure 2: Daily variation around the average number of tweets discussing ISIS
(variable name: ATTENTION DEVIATION)*

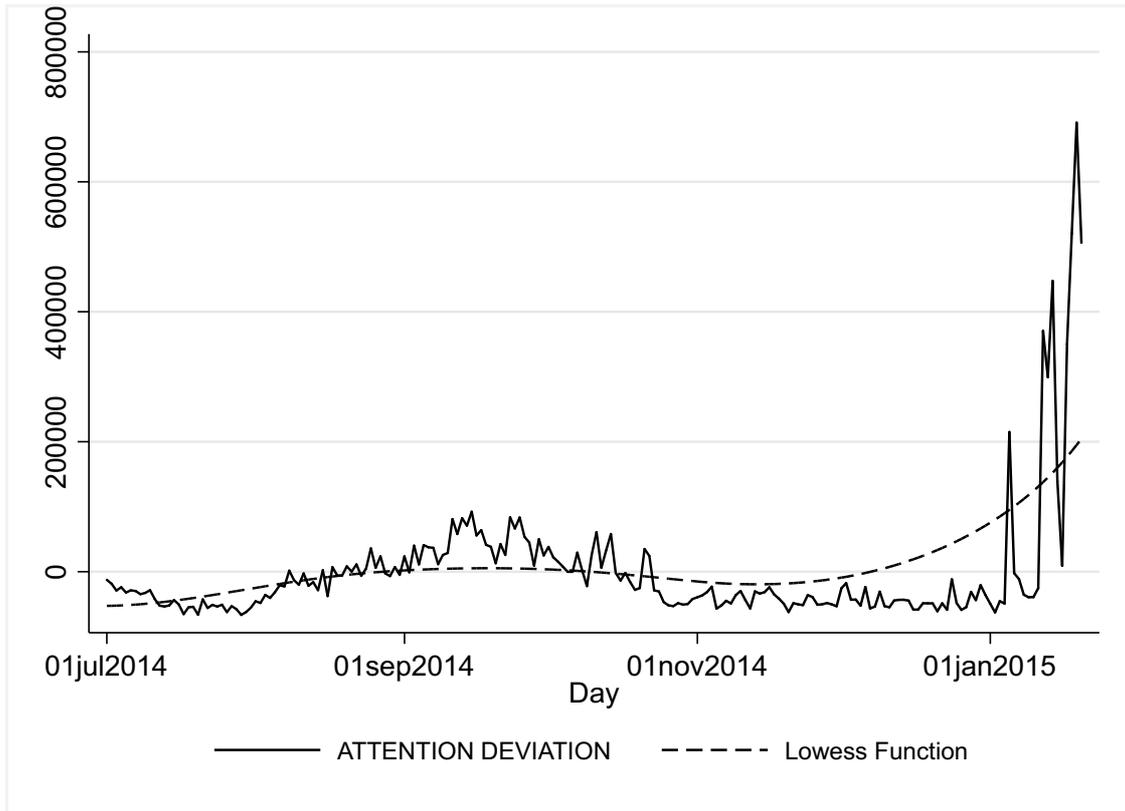

*Figure 3: Predicted number of Foreign Fighters for different values of* Sentiment toward ISIS

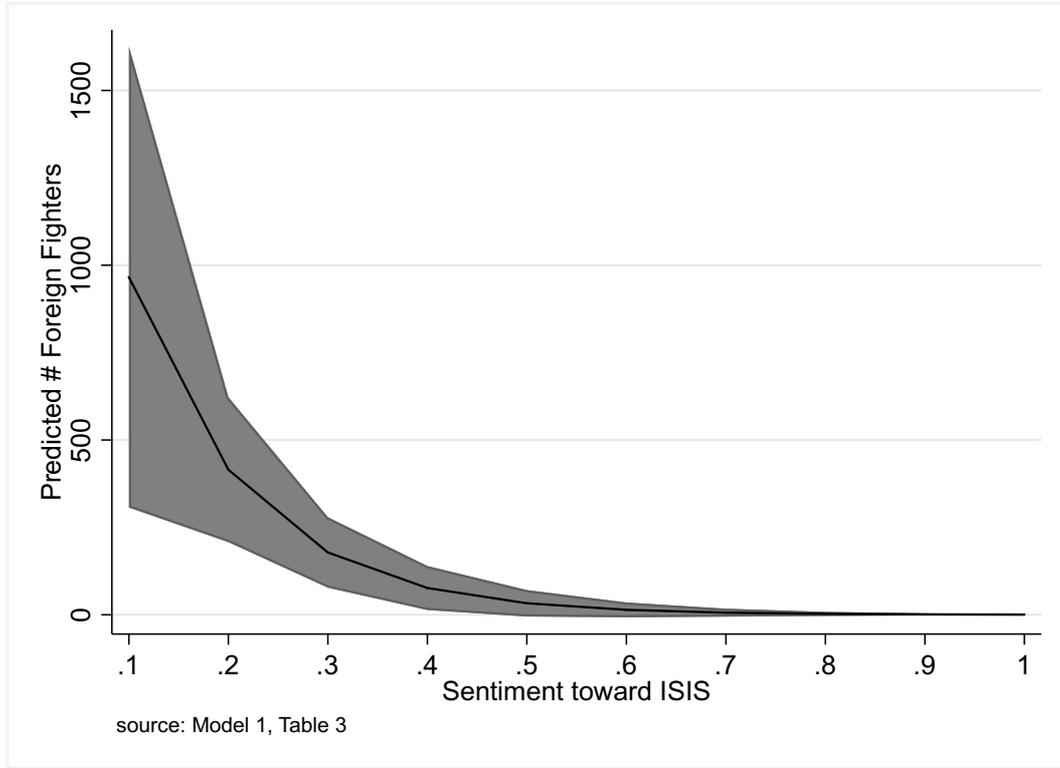

**Tables**

*Table 1: Examples of classified tweets about ISIS in the training-set*

| *Positive classification* | *Negative classification* |
|---|---|
| 17 July 2014<br>يحاربون أنهم ادركتم هل الجزية وفرض الرجم حد يهاجمون ولةالد تطبقه لذيا الله شرع يكرهون لانهم الإسلامية الدولة زعموا كما وليس<br>(English translation: "They are attacking the rules on stoning and taxing non-believers. They are fighting the Islamic State because they hate god's shariah that Islamic State implements") | 18 August 2014<br>من ليس معنا فهو ضدنا" كلمات تلفظها القرد بغدادي #الدولة_الإسلامية #باغيةنبوش وطبقها همج الـ<br>(English translation: Either you're with us or against us. These words were said by that monkey Bush and are put into effect by [Abu Bakr all]Bagdhadi's mobs. #unjust #Islamicstate) |
| 29 August 2014<br>ولكة على داعش ولا بوش لك أقرب من الدوله جنود تكره ليش ماعتقد<br>شيعهال هم الدوله يكره من لان قهالطريه بهذه وتتكلم سني إنك<br>(English translation: "Why do you hate Islamic State soldiers? Who are you closer to Bush or Isis? According to what you are saying I don't | 29 September 2014<br>مش قتل طفل سوري عشانه سرق ! الله ياخذهم ويبيدهداء<br>): المنظر بشع ما قدرت أشوفه زين ،، من الأرض<br>(English translation: Isis killed a syrian boy for stealing. May god take them and wipe them off the face of the earth. The images are horrible, I couldn't watch it okay :( |

- 31 -

| | |
|---|---|
| believe you are Sunni and you talk in this way beacuse those who hate Islamic State are Shia") | |
| 7 September 2014<br>نيحبو هم .. معك يختلفون وسوريا العراق في السنة أهل<br>يف وتعيش الدولة تكره أنتو معها وعاشوا الإسلامية الدولة<br>قطر ..<br>(English translation: The Sunni people in Iraq and Syria disagree with you. They live within Islamic State and they love the state but you hate Isis though you live in Qatar) | 14 October 2014<br>جاهل من يقبل او يؤيد فكر داعش الدموي الحيواني المتطرف بالدين الاسلامي<br>مل لحساب اعداء الاسلاموسماحته ..ومتأكد ان داعش تع<br>(English translation: Those who call themselves Isis and Islamic State, keep quite cos a superior complex is not good and your path is ignorance and darkness. Educate yourself, it's better than being a vagabond from the streets) |
| 22 October 2014<br>..أجمل ما في الشام<br>ذبساطة كأنك لم تعد بحاجة لتوعي العوام، فهم قد فهموا بد الإعلام بالحديث عن #داعش<br>(English translation: The most beautiful thing in Syria is that you don't need to reaise the consciouness of the people because they have understood the lies of the media when they talk about Isis) | 17 Octotber 2014<br>...أكره داعش وكل من سار في فلكها أنا<br>(وانت ماذا تنتظر) بنَ...<br>(English translation: I hate Isis and anyone moving in its orbit. And what are you waiting for?) |

*Table 2: Correlates of daily* SENTIMENT DEVIATION

| | Model 1 | Model 2 | Model 3 |
|---|---|---|---|
| SENTIMENT DEVIATION Lag | 0.091 | 0.074 | 0.049 |
|  | (0.096) | (0.095) | (0.099) |
| Attacks v/ Mosque/Imam | -0.046* | -0.054** | -0.048** |
|  | (0.019) | (0.020) | (0.018) |
| ISIS Military victory) | -0.047** | -0.049** | -0.048** |
|  | (0.017) | (0.017) | (0.016) |
| ISIS Military defeat | -0.007 | -0.008 | -0.008 |
|  | (0.024) | (0.023) | (0.023) |
| Beheadings (Western) | 0.116 | 0.121 | 0.109 |
|  | (0.091) | (0.093) | (0.091) |
| Beheadings (Non-Western | -0.049 | -0.049 | -0.049 |
|  | (0.033) | (0.034) | (0.031) |
| 'Muslims Unity' ISIS speeches | -0.043* | -0.042*** | -0.048** |
|  | (0.017) | (0.012) | (0.015) |
| ATTENTION DEVIATION (/10000) | - | -0.002*** | -0.002*** |
|  |  | (0.001) | (0.001) |
| Charlie Hebdo attack | - | -0.117*** | -0.111*** |
|  | (0.027) | (0.028) |  |
| News on-line Lag | - | - | 0.000* |
|  |  |  | (0.000) |
| Constant | 0.008 | 0.009 | -0.037 |
|  | (0.009) | (0.009) | (0.024) |
| *N* | 203 | 203 | 203 |
| *BIC* | -301.973 | -306.219 | -305.138 |

Robust standard errors in parentheses
[+] $p < 0.10$, [*] $p < 0.05$, [**] $p < 0.01$, [***] $p < 0.001$

*Table 3: Explaining the Number of Foreign Fighters by country*

|  | (1) Model 1 | (2) Model 2 | (3) Model 3 | (4) Model 4 | (5) Model 5 |
|---|---|---|---|---|---|
| - Sentiment toward ISIS | -8.451*** | -22.36* | -8.387*** | -8.787*** | - |
|  | (1.933) | (11.18) | (1.966) | (2.043) |  |
| - Justify Attacks | - | - | - | - | -0.108*** |
|  |  |  |  |  | (0.0318) |
| - Active Islamic Terrorist Group within a country | 1.224 | 0.282 | 1.288[+] | 0.646 | - |
|  | (0.754) | (0.805) | (0.747) | (0.752) |  |
| - Living in a country around ISIS border | 1.299** | 1.892* | 1.322** | 0.367 | - |
|  | (0.488) | (0.868) | (0.465) | (0.635) |  |
| - % Shia over Muslims within a country | -0.0741*** | -0.0694*** | -0.0779*** | -0.0143 | - |
| | (0.0143) | (0.0179) | (0.0134) | (0.0183) | |
| - Democracy score (PolityIV) | -0.0694 | -0.121 | -0.0797$^+$ | -0.118*** | - |
| | (0.0435) | (0.0904) | (0.0435) | (0.0347) | |
| - % Broadband | 0.129*** | 0.181** | 0.125*** | 0.119*** | - |
| | (0.0342) | (0.0605) | (0.0328) | (0.0267) | |
| Constant | 2.610** | 4.935* | 2.626** | 3.504*** | 4.979*** |
| | (0.845) | (2.081) | (0.849) | (0.809) | (0.612) |
| | (0.227) | (0.336) | (0.233) | (0.261) | |
| Observations | 61 | 23 | 60 | 52 | 22 |
| Log Pseudolikelihood | -297.5 | -120.5 | -290.1 | -286.1 | -153.2 |

Robust standard errors in parentheses
$^+ p < 0.10$, $^* p < 0.05$, $^{**} p < 0.01$, $^{***} p < 0.001$

*Note: The dependent variable is the number of Foreign Fighters from a given country. Entries are negative binomial coefficients with exposure term = ln(percentage of Muslim people living in a given country. Source: https://en.wikipedia.org/wiki/Islam_by_country).*

**Appendix: List of keyword employed in our query**

الزرقاوي مسعد ابو OR الشوري لمجلس مجاهدين OR الجهاد و التوحيد منظمه OR الشام و العراق في الاسلامية الدولة و العراق في القاعدة عن الانفصاليين OR خليفه OR البغدادي بكر ابو خليفه OR عالمية جهاد حركه OR الخوري ليث OR الإسلامية الدولة OR ISIS OR ISIL OR داعش OR الإسلامية الدولة OR لقاعدها OR الاسلامية الدوله OR سوريا وتتمدد باقية OR والشام العراق في الإسلامية الدولة OR داعش OR والشام العراق في الإسلامية الدولة